\documentclass[11pt]{article}

\usepackage{acl}

\usepackage{times}
\usepackage{latexsym}
\usepackage{amsmath,amssymb}
\usepackage{listings}
\usepackage[disable]{todonotes}
\usepackage{booktabs}
\usepackage{multirow}
\usepackage{multicol}
\usepackage[T1]{fontenc}
\newcommand{\evaluate}[2]{\left\llbracket #1 \right\rrbracket_{#2}}
\newcommand{\fv}{\mathrm{fv}}
\usepackage{stmaryrd}
\usepackage{xspace}
\newcommand{\sysname}{TL;QR\xspace}

\usepackage[utf8]{inputenc}
\newcommand{\Domain}{\mathcal{D}}

\usepackage{microtype}

\usepackage{inconsolata}

\usepackage{graphicx}
\usepackage{pgfplots}
\usepackage{tikz}
\usetikzlibrary{positioning,arrows.meta,calc,fit,backgrounds}
\definecolor{poscol}{rgb}{0.13,0.33,0.72}
\definecolor{negcol}{rgb}{0.75,0.13,0.13}
\definecolor{shiftcol}{rgb}{0.85,0.45,0.0}
\definecolor{stageblue}{rgb}{0.20,0.35,0.60}
\definecolor{regexcol}{rgb}{0.0,0.45,0.15}

\title{TL;QR: Querying Temporal Discursive Patterns in Textual Corpora}

\author{
  \textbf{François Hublet\textsuperscript{1}},
  \textbf{Vladimir Krsmanović\textsuperscript{2}},
  \textbf{Jesús Solano\textsuperscript{1}},
  \\
  \textbf{Ahmed Bouhoula\textsuperscript{1}},
  \textbf{Ana Oliveira da Costa\textsuperscript{3}}
  \\
  \\
  \textsuperscript{1}\textit{ETH Zürich},
  \textsuperscript{2}\textit{CISPA Helmholtz Center for Information Security},
  \textsuperscript{3}\textit{IST Austria},
  \\
  \small{
    \textbf{Correspondence:} \href{mailto:francois.hublet@inf.ethz.ch}{francois.hublet@inf.ethz.ch}
  }
}

\definecolor{red}{rgb}{0.85,0.0,0.0}   % lighter red
\definecolor{green}{rgb}{0.0,0.5,0.0}  % darker-ish green (readable)
\definecolor{blue}{rgb}{0.0,0.0,0.7}   % lighter blue

\newcommand{\kw}[1]{{\color{red}\mathtt{#1}}}
\newcommand{\Once}{\mathbin{\kw{ONCE}}}
\newcommand{\Since}{\mathbin{\kw{SINCE}}}
\newcommand{\Prev}{\mathop{\kw{PREV}}}
\newcommand{\agg}[5]{#1 \gets \kw{#2}(#3; #4; #5)}
\renewcommand{\And}{\mathbin{\kw{AND}}}
\newcommand{\Or}{\mathbin{\kw{OR}}}
\newcommand{\Not}{\mathop{\kw{NOT}}}
\newcommand{\Exists}{{\mathop{\kw{EXISTS}}}}
\newcommand{\Let}{\mathop{\kw{LET}}}
\newcommand{\In}{\mathop{\kw{IN}}}
\newcommand{\prompt}[1]{{\color{blue}``\mathtt{#1}"}}
\newcommand{\regexp}[2]{{\color{green}/\mathtt{#1}/}#2}

\lstdefinelanguage{MFOTL-NL}{
  sensitive=true,
  morekeywords={
    ONCE,SINCE,PREV,AND,OR,NOT,EXISTS,LET,IN,CNT,tp
  },
  keywordstyle=\color{red}\ttfamily,
  basicstyle=\scriptsize\ttfamily,
  morecomment=[l]{//},
  commentstyle=\itshape\color{gray},
  morestring=[b]",            % double-quoted strings
  stringstyle=\color{blue},
  showstringspaces=false,
  moredelim=[s][\color{green}\ttfamily]{/}{/},
  escapechar=@,
}

\begin{document}
\maketitle
\begin{abstract}
Many questions about large textual corpora concern patterns that unfold over time, such as a political group whose stance on a topic drifts from support to criticism over several years of debates.
We present \sysname (\emph{Too Long; Query with logical temporal Reasoning}), a declarative query language and open-source engine for finding such temporal discursive patterns.
In \sysname, a single query combines natural-language predicates, whose truth is judged by a Large Language Model (LLM), with regular expressions and the operators of a temporal logic.
The LLM answers only local questions about individual text passages; a symbolic engine then aggregates these judgments over time and links every match to the recorded passages and predicate decisions behind it.
With this division of labor, the temporal and relational reasoning stays exact and auditable even over long time horizons.
In addition, when a query includes a regular-expression pre-filter, only a small fraction of the corpus ever reaches the LLM.
On a suite of controlled synthetic benchmarks with temporally and semantically complex patterns, \sysname is substantially more accurate than LLM-only baselines.
In a case study on ten years of European Parliament debates (442{,}593 multilingual speeches), \sysname identifies candidate shifts in the political groups' stances on the GDPR privacy regulation.
\end{abstract}

\section{Introduction}
Analyzing large textual corpora for complex discursive phenomena, such as stance shifts, recurring rhetorical structures, or argumentative patterns, is central to computational social science, political science, and the digital humanities. 
Consider a question that a political scientist may ask about a multi-year parliamentary record: \emph{when, if ever, did a political group's stance on environmental regulation flip from mostly supportive to mostly critical?} Answering it requires three abilities at once: understanding \emph{what} each
speech expresses (``this speech criticizes environmental regulation''), tracking \emph{when} speeches occur and in what order (``after months of mostly supportive speeches''), and relating \emph{who} produced them (``speeches by members of the same group'').

Since these questions involve open-ended language understanding, LLMs are a natural starting point as they can readily automate many annotation tasks~\cite{DBLP:journals/corr/abs-2303-15056,Ziems2024}. The above analysis, however, demands more than just understanding each text in isolation. It also requires reasoning over time and over the relations between texts at different time-points. Annotations must be counted, ordered, and compared across speakers and over months or years. Unfortunately, LLMs are known to be unreliable on such long-horizon tasks~\cite{DBLP:journals/corr/abs-2404-06654,DBLP:conf/emnlp/GaoLYBK24,DBLP:conf/acl/ChuCCY00024,DBLP:conf/iclr/FatemiKTMYPSHP25}, and processing entire corpora with them can be costly. In contrast, traditional corpus query tools~\cite{Hardie2012,Krause2016} can transparently and cheaply aggregate annotations but cannot interpret open-ended meaning.

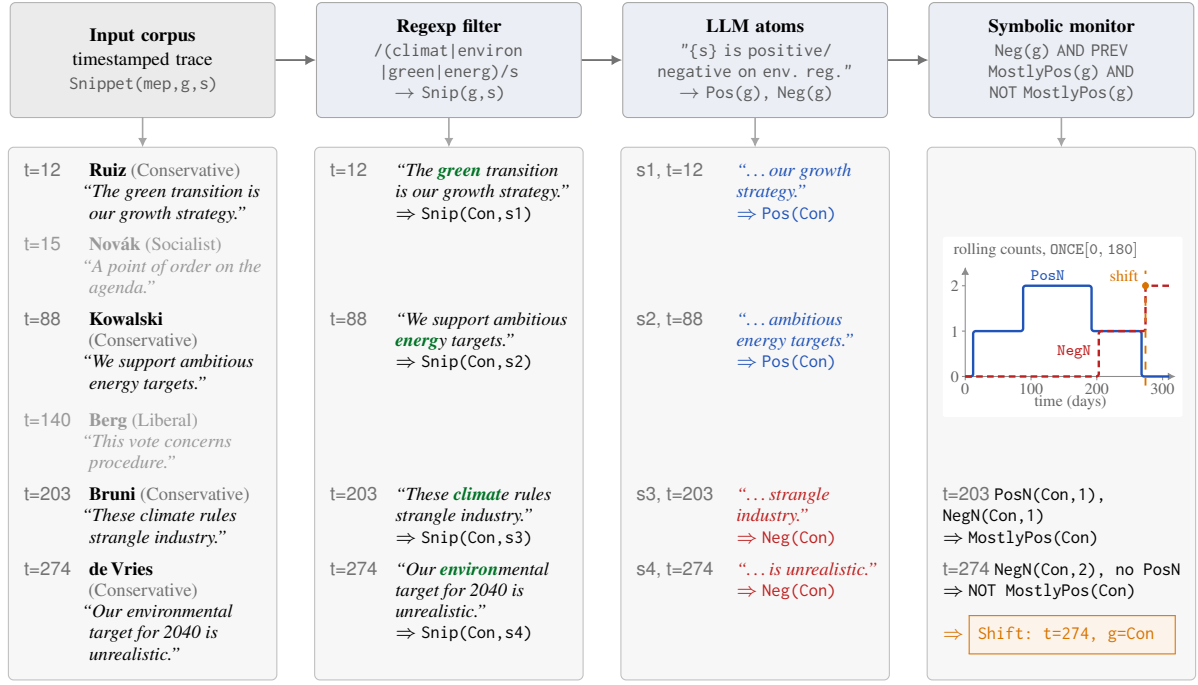
\begin{figure*}[thb]
\centering
\hyphenpenalty=10000 \exhyphenpenalty=10000

\newcommand{\lbl}[1]{\strut\color{black!55}\sffamily #1}
\newcommand{\lblg}[1]{\strut\color{black!40}\sffamily #1}
% #1 speaker, #2 group, #3 text
\newcommand{\evc}[3]{\raggedright\strut\mbox{\textbf{#1}}
  \mbox{{\color{black!55}(#2)}}\\ \emph{``#3''}}
\newcommand{\evcg}[3]{\raggedright\strut\color{black!40}\mbox{\textbf{#1}}
  \mbox{(#2)}\\ \emph{``#3''}}
% #1 short text, #2 derived predicate
\newcommand{\derc}[2]{\raggedright\strut\emph{``#1''}\\
  $\Rightarrow$~{\ttfamily #2}}
% #1 colour, #2 short text, #3 derived predicate
\newcommand{\dercc}[3]{\raggedright\strut\color{#1}\emph{``#2''}\\
  $\Rightarrow$~{\ttfamily #3}}
% #1 regexp-matching substring
\newcommand{\m}[1]{{\color{regexcol}\bfseries #1}}
% #1 time-point, #2 counts derived there, #3 predicate they justify
\newcommand{\derm}[3]{\raggedright\strut{\color{black!55}\sffamily #1}~%
  {\ttfamily #2}\\ $\Rightarrow$~{\ttfamily #3}}
% rolling-count plot shown in the monitor panel
\newcommand{\plot}{%
\begin{tikzpicture}[font=\tiny, baseline=0pt, x=0.87cm]
  \node[anchor=south west, black!60, inner sep=0pt] at (-0.18,1.56)
     {rolling counts, $\mathtt{ONCE[0,180]}$};
  \draw[black!45, -{Latex[length=3pt,width=3pt]}] (0,0) -- (3.20,0);
  \draw[black!45, -{Latex[length=3pt,width=3pt]}] (0,0) -- (0,1.45);
  \foreach \x/\l in {0/0, 1/100, 2/200, 3/300}
     \draw[black!45] (\x,0) -- (\x,-0.05) node[below, black!60, inner sep=1pt]{\l};
  \foreach \y/\l in {0/0, 0.6/1, 1.2/2}
     \draw[black!45] (0,\y) -- (-0.05,\y) node[left, black!60, inner sep=1pt]{\l};
  \node[black!60, anchor=north, inner sep=1pt] at (1.6,-0.20) {time (days)};
  \draw[poscol, line width=0.9pt]
     (0,0) -- (0.12,0) -- (0.12,0.6) -- (0.88,0.6) -- (0.88,1.2)
     -- (1.92,1.2) -- (1.92,0.6) -- (2.68,0.6) -- (2.68,0) -- (3.10,0);
  \draw[negcol, line width=0.9pt, dash pattern=on 2.5pt off 1.6pt]
     (0,0) -- (2.03,0) -- (2.03,0.6) -- (2.74,0.6) -- (2.74,1.2) -- (3.10,1.2);
  \draw[shiftcol, dashed, line width=0.7pt] (2.74,-0.12) -- (2.74,1.40);
  \fill[shiftcol] (2.74,1.2) circle (1.3pt);
  \node[shiftcol, anchor=south east, inner sep=1.5pt] at (2.70,1.20) {shift};
  \node[poscol, anchor=west, inner sep=1pt] at (0.95,1.34) {$\mathtt{PosN}$};
  \node[negcol, anchor=north west, inner sep=1pt] at (1.35,0.45) {$\mathtt{NegN}$};
\end{tikzpicture}}
\begin{tikzpicture}[
  font=\scriptsize,
  panel/.style={draw=black!25, rounded corners=2pt, fill=black!3},
  head/.style={draw=black!35, rounded corners=2pt, fill=stageblue!10,
               text width=3.15cm, inner sep=5pt, anchor=north, align=center,
               minimum height=1.52cm},
  inhead/.style={head, fill=black!8, draw=black!30},
  flow/.style={-{Latex[length=4pt,width=4pt]}, black!55, line width=0.7pt},
  drop/.style={-{Latex[length=3.5pt,width=3.5pt]}, black!35, line width=0.6pt},
]

%% ------------------------------------------------------------------ headers
\node[inhead] (h1) at (0,0)
  {\textbf{Input corpus}\\[1pt]
   {\scriptsize timestamped trace}\\[1pt]
   {\ttfamily\color{black!65}Snippet(mep,g,s)}};
\node[head] (h2) at (4.05,0)
  {\textbf{Regexp filter}\\[1pt]
   {\ttfamily\color{black!65}/(climat|environ\\|green|energ)/s\\ $\to$ Snip(g,s)}};
\node[head] (h3) at (8.10,0)
  {\textbf{LLM atoms}\\[1pt]
   {\ttfamily\color{black!65}"\{s\} is positive/\\negative on env.\ reg."\\ $\to$ Pos(g), Neg(g)}};
\node[head] (h4) at (12.15,0)
  {\textbf{Symbolic monitor}\\[1pt]
   {\ttfamily\color{black!65}Neg(g) AND PREV\\MostlyPos(g) AND\\NOT MostlyPos(g)}};

\draw[flow] (h1.east) -- (h2.west);
\draw[flow] (h2.east) -- (h3.west);
\draw[flow] (h3.east) -- (h4.west);

%% --------------------------------------------- panels 1--3: one shared grid
\node[anchor=north west, inner sep=0pt] (grid) at (-1.575,-2.096) {%
\begin{tabular}{@{}p{0.86cm}@{}p{2.29cm}@{\hspace{0.90cm}}p{0.86cm}@{}p{2.29cm}@{\hspace{0.90cm}}p{1.32cm}@{}p{1.83cm}@{\hspace{0.90cm}}p{0.86cm}@{}p{2.29cm}@{}}
  \lbl{t=12} & \evc{Ruiz}{Conservative}{The green transition is our growth strategy.} &
  \lbl{t=12} & \derc{The \m{green} transition is our growth strategy.}{Snip(Con,s1)} &
  \lbl{s1, t=12} & \dercc{poscol}{\dots our growth strategy.}{Pos(Con)} &
  &
  \tabularnewline[20pt]
  \lblg{t=15} & \evcg{Nov\'ak}{Socialist}{A point of order on the agenda.} &
  & & & & &
  \tabularnewline[20pt]
  \lbl{t=88} & \evc{Kowalski}{Conservative}{We support ambitious energy targets.} &
  \lbl{t=88} & \derc{We support ambitious \m{energ}y targets.}{Snip(Con,s2)} &
  \lbl{s2, t=88} & \dercc{poscol}{\dots ambitious energy targets.}{Pos(Con)} & &
  \tabularnewline[30pt]
   \lblg{t=140} & \evcg{Berg}{Liberal}{This vote concerns procedure.} &
  & & & & &
  \tabularnewline[20pt]
  \lbl{t=203} & \evc{Bruni}{Conservative}{These climate rules strangle industry.} &
  \lbl{t=203} & \derc{These \m{climat}e rules strangle industry.}{Snip(Con,s3)} &
  \lbl{s3, t=203} & \dercc{negcol}{\dots strangle industry.}{Neg(Con)} &
  \multicolumn{2}{@{}p{3.15cm}@{}}{\derm{t=203}{PosN(Con,1), NegN(Con,1)}{MostlyPos(Con)}}
  \tabularnewline[20pt]
  \lbl{t=274} & \evc{de\,Vries}{Conservative}{Our environmental target for 2040 is unrealistic.} &
  \lbl{t=274} & \derc{Our \m{environ}mental target for 2040 is unrealistic.}{Snip(Con,s4)} &
  \lbl{s4, t=274} & \dercc{negcol}{\dots is unrealistic.}{Neg(Con)} &
  \multicolumn{2}{@{}p{3.15cm}@{}}{\derm{t=274}{NegN(Con,2), no PosN}{NOT MostlyPos(Con)}\\[5pt]
    {\color{shiftcol}$\Rightarrow$}~\fcolorbox{shiftcol}{shiftcol!8}{%
      \parbox{2.50cm}{\raggedright\color{shiftcol}\strut
      {\ttfamily Shift: t=274, g=Con}\strut}}}
  \tabularnewline
\end{tabular}};

%% the rolling-count plot sits at the top of the monitor panel
\node[anchor=north west, inner sep=0pt, fill=white, rounded corners=1pt, inner sep=2pt] at (10.575,-3.096) {\plot};

%% ------------------------------------------------- panel frames + bottom row
\path (grid.south) ++(0,-0.176) coordinate (bot);
\begin{scope}[on background layer]
  \foreach \c in {0, 4.05, 8.10, 12.15}
    \draw[panel] ({\c-1.775},-1.92) rectangle ({\c+1.775},0 |- bot);
\end{scope}

\foreach \c in {0, 4.05, 8.10, 12.15}
  \draw[drop] (\c,-1.52) -- (\c,-1.92);

\end{tikzpicture}
\caption{The \sysname pipeline, illustrated on the query of
Figure~\ref{fig:query} over a hypothetical corpus with three groups. Each
panel shows the intermediate predicate it generates: the regexp pre-filter
turns on-topic speeches into $\mathtt{Snip}$ events, the LLM atoms label the
resulting snippets {\color{poscol}positive} or {\color{negcol}negative} and
emit $\mathtt{Pos}/\mathtt{Neg}$, and the symbolic monitor derives the rolling
counts $\mathtt{PosN}/\mathtt{NegN}$ and $\mathtt{MostlyPos}$, and returns the
{\color{shiftcol}detected stance shift}. Rows are
aligned across the first three panels, so a gap marks a speech the filter
dropped. Only the LLM-atoms panel calls a model.}
\label{fig:pipeline}
\end{figure*}

In this paper, we present \sysname (\emph{Too Long; Query with logical
temporal Reasoning}), a neuro-symbolic tool that combines the precision of temporal-logic reasoning with the versatility of natural-language annotations. The user states the pattern of interest as a single declarative query in a formal language, using operators for filtering, classification, counting, comparison, as well as natural language and regular expression patterns. An LLM answers only local questions about individual text fragments, such as ``does this passage support environmental regulation?'', while a symbolic
engine with a formally defined semantics combines these local answers across time, performing all counting, ordering, and windowing operations.
Figure~\ref{fig:pipeline} shows this data flow on our running example: a cheap regular-expression filter selects candidate speeches, LLM-backed predicates test each candidate for a supportive and critical stance, and the symbolic engine detects the group whose stance shifts. 

The core idea of \sysname is grounding natural language into atoms in Metric First-Order Temporal Logic
(MFOTL)~\cite{Chomicki1995,Basin2015,Basin2015b}.
MFOTL acts as the semantic foundation for our specification language, providing operators for quantification, comparisons, and boolean and interval operations. In \sysname, the atomic predicates of this logic may be corpus events, regular expressions, or natural-language descriptions whose truth is decided by an LLM backend. 

To the best of our knowledge, \sysname is the first query language for text that combines LLM-judged atomic predicates with metric temporal operators, first-order quantification, and aggregations under a formal semantics. The formal semantics gives each query a single unambiguous meaning, avoiding misunderstandings that arise when a pattern is specified in ad hoc code, such as Python, or in imprecise natural language. Furthermore, the use of formally defined semantics and the symbolic reasoning engine provides auditability without requiring additional machinery or external mechanisms. Every verdict is derived from recorded predicate-level decisions, so each reported match comes with concrete passages and counts, and the user can always trace a verdict back to the annotations used to decide it.

Our contributions are:
\begin{enumerate}
    \item \sysname, a specification language that integrates MFOTL with
    regular-expression and LLM-backed predicates, together with an
    open-source engine that compiles queries into a hybrid neuro-symbolic
    monitoring problems and evaluates them incrementally over timestamped
    corpora (Section~\ref{sec:approach}).
    \item A synthetic benchmark suite of temporally and semantically complex
    discursive phenomena, which provides a controlled testbed for the
    long-horizon reasoning of LLM pipelines.
    \sysname outperforms LLM-only
    baselines on all benchmarks (Section~\ref{sec:eval}).
    \item A case study on ten years of European Parliament debates
    ($442{,}593$ multilingual speeches) demonstrating the scalability of our approach. \sysname identifies
    candidate shifts in the political groups' stances on the GDPR for under
    \$2 of LLM cost; a prompt-only pipeline over the same corpus would cost
    orders of magnitude more without providing the same level of transparency and reliability (Section~\ref{sec:case}).
\end{enumerate}

\section{Related Work}\label{sec:related}

\paragraph{LLMs for annotation and aggregation.}
LLMs are increasingly used as zero- and few-shot annotators for open-ended text analysis, semantic extraction, and annotation in computational social science~\cite{Ziems2024}. Their strong performance on these tasks, however, does not carry over to reasoning over long time horizons, e.g., counting and
ordering~\cite{DBLP:conf/emnlp/GaoLYBK24,DBLP:conf/acl/ChuCCY00024}. This contrast motivates our design, which uses LLMs only as annotators, while long-horizon reasoning is handled by a symbolic tool. %, and positions our work within a broader line of research that couples LLMs with symbolic structures to improve temporal reasoning or question-answering performance.

\paragraph{Corpus query languages.}
For the analysis of large text corpora, CQPweb~\cite{Hardie2012} and
ANNIS~\cite{Krause2016} provide scalable and inspectable query systems with user-provided annotations. They support expressive lexical and structural patterns but require manual annotation and cannot process natural-language meanings. Declarative information-extraction and weak-supervision systems such as SystemT~\cite{li2011systemt}, DeepDive~\cite{shin2015incremental}, and Snorkel~\cite{ratner2017snorkel} instead help construct structured annotations or supervision for downstream analysis. More recent semantic data-processing systems incorporate LLMs directly into relational or dataflow pipelines. LOTUS~\cite{patel2024semantic}, Palimpzest~\cite{liu2025palimpzest}, and DocETL~\cite{shankar2024docetl} support operations such as semantic filtering, mapping, joining, grouping, and aggregation. As these systems target batch or record-wise processing, they do not support temporal reasoning. \sysname provides a more expressive specification language, temporal reasoning, and direct support for LLM-judged atoms inside its formulae.

\paragraph{Temporal pattern detection over unstructured streams.}
Detecting patterns over time and over streams of data is the focus of
event processing research. In computer vision, neuro-symbolic video
search~\cite{DBLP:journals/corr/nsvs,DBLP:conf/aaai/ShahSGCMPMC26} follows a similar extract-then-reason pattern over frame sequences. Closest to our work is VectraFlow~\cite{chen2026vectraflow}, which combines LLM-based extraction from document streams with automata for matching. Its pattern operator supports sequences, conjunction, disjunction, repetition, bounded negation, and temporal windows, and its broader dataflow system provides semantic joins, grouping, and aggregation, similar to \sysname. Unlike these approaches, \sysname is grounded in MFOTL, a formal language with a well-defined syntax and semantics that has been used extensively for trace-based monitoring of industrial systems~\cite{Basin2023}. 

\paragraph{LLMs and symbolic reasoning.} Neuro-symbolic research combines neural and symbolic methods to build more robust and powerful systems. Most work in this direction either translates natural language into specifications~\cite{DBLP:conf/emnlp/ChenGZF23,DBLP:conf/cav/CoslerHMST23} or attaches symbolic solvers to improve LLM reasoning~\cite{DBLP:conf/emnlp/PanAWW23,DBLP:conf/emnlp/OlaussonGLZSTL23}. Other approaches, such as Binder~\cite{cheng2022binding}, Faithful-CoT~\cite{lyu2023faithful}, and PAL~\cite{gao2022pal}, use LLMs to generate
programs and then execute them. We use LLMs differently: in \sysname they
\emph{evaluate predicates} over natural text that the symbolic engine
then reasons over, rather than generating programs.

\section{\sysname: Language and Engine} % approach
\label{sec:approach}

\sysname consists of two components: a declarative \emph{query language}
(Section~\ref{sec:lang}) in which the user states the pattern of interest,
and a \emph{query engine} (Section~\ref{sec:engine}) that evaluates the query efficiently over a large corpus. We introduce both through the running example from the introduction, which we denote by $(\ast)$: \emph{over a multi-year parliamentary corpus, report each political group whose stance on environmental regulation flips from mostly positive to mostly negative, where ``mostly'' is measured over a rolling $180$-day window.} Note that nothing in our design is specific to parliamentary corpora or stance detection. A query may range over any timestamped collection of texts (social-media streams, forum posts, clinical notes, etc.), and its atoms may encode any annotation task, from topical relevance to rhetorical function. For instance, the benchmarks in Section~\ref{sec:eval} use the same constructs to detect speakers who evade commitments or coordinate hostile campaigns. 

\subsection{Query Language}\label{sec:lang}

The \sysname query language extends past-only Metric First-Order Temporal Logic (MFOTL)~\cite{Chomicki1995} with two new kinds of atomic predicates: natural-language prompts and regular-expression (regexp) predicates. To show how the query language works in practice, a complete query for $(\ast)$ is shown in
Figure~\ref{fig:query}, and the full syntax in
Figure~\ref{fig:query-language}. The formal semantics follows \citet{Hublet2025} and is given in
Appendix~\ref{sec:semantics}. 

\begin{figure}
\begin{lstlisting}[language=MFOTL-NL]
LET Snip(g, s) = EXISTS mep. Snippet(mep, g, s)
  AND /(climat|environ|green|energ)/s IN
LET Pos(g) = EXISTS s. Snip(g, s)
  AND "{s} is positive on
       environmental regulation" IN
LET Neg(g) = EXISTS s. Snip(g, s)
  AND "{s} is negative on
       environmental regulation" IN
LET PosN(g,x) = x <-
  CNT(t; g; ONCE[0,180] (Pos(g) AND tp(t))) IN
LET NegN(g,y) = y <-
  CNT(t; g; ONCE[0,180] (Neg(g) AND tp(t))) IN
LET MostlyPos(g) = EXISTS x,y.
  PosN(g,x) AND NegN(g,y) AND x >= y IN
Neg(g) AND PREV MostlyPos(g) AND NOT MostlyPos(g)
\end{lstlisting}
\caption{\sysname query capturing the running example $(\ast)$, read
top-down as a cascade: regexp filter ($\mathtt{Snip}$), LLM stance atoms
($\mathtt{Pos}$/$\mathtt{Neg}$), rolling counts
($\mathtt{PosN}$/$\mathtt{NegN}$), majority comparison
($\mathtt{MostlyPos}$), and the shift condition (last
line).\label{fig:query}}
\end{figure}

\paragraph{Natural-language predicates.} The most distinctive atoms of \sysname are natural-language predicates: quoted sentences containing placeholders of the form $\prompt{\{x\}}$ that stand for string variables. At evaluation time, an automatic judge (i.e., an LLM backend) decides whether the instantiated sentence is true. In $(\ast)$, the atom {\color{blue}\texttt{"\{s\} is positive on environmental regulation"}} holds for a speech snippet $s$ if and only if the LLM judges this claim to be true of $s$. Such atoms let queries refer to open-ended semantic phenomena (stance,
topical relevance, rhetorical figures, etc.) without any prior annotation of the corpus.

\paragraph{Regexp predicates.} For surface-level matching that does not require a model call, \sysname supports regexp predicates of the form $\regexp{\mathit{r}}{x}$, where $r$ is a Python-style regular expression matched against the string variable $x$. For example, in $(\ast)$, the atom $\regexp{(climat|environ|green|energ)}{s}$ keeps only snippets that mention the topic at all. In this notation, the trailing $s$ denotes the string variable being matched. The core benefit of regexp predicates is their low computational cost, which  makes them a natural complement to the natural-language predicates. This combination enables uses where Regexp predicates can be used to filter patterns and pass the content to the LLM only if the content matches the regex query, avoiding unnecesary token costs.

%While this greatly reduced the number of LLM calls required it is a trade-off between recall performance and token consumption, 
%as filtered-out snippets that discuss the topic could be filtered out of the regex queries are overly resitrictive. For example, to avoid this the case study in Section~\ref{sec:case} uses a $42$-term multilingual alternation) to cover possible . For phenomena without a reliable surface signature, the pre-filter can simply be omitted, at the price of one LLM call per candidate snippet, as in our benchmarks (Section~\ref{sec:eval}).

\paragraph{Data model.} Before composing these atoms, we fix how the corpus is represented. A corpus is a \emph{trace}: a sequence of timestamped \emph{databases} $(\tau_1,D_1),\dots,(\tau_n,D_n),\dots$ with $\tau_1 \leq \tau_2 \leq \cdots$, where each $D_i$ is a finite set of \emph{events} $\mathtt{e}(d_1,\dots,d_k)$ and the index $i$ is called a \emph{time-point}. In our example, every speech snippet contributes an event $\mathtt{Snippet}(mep, g, s)$ recording the speaker, their group $g$, and the snippet text $s$. A query $\phi$ is evaluated at each time-point $i$ and returns all valuations $v$ of its free variables satisfying a sequent $v, i \models \phi$,
reading ``the variable assignment $v$ satisfies $\phi$ at time-point $i$''~\cite{Basin2015}. In $(\ast)$, the only free variable is the group $g$, so each returned valuation names a group whose stance has just shifted.

\paragraph{First-order operators.} The Boolean connectives $\kw{AND}$, $\kw{OR}$, $\kw{NOT}$ and the existential quantifier $\kw{EXISTS}$ combine atoms, as in first-order logic or in SQL. Built-in arithmetic symbols ($\texttt{+}$, $\texttt{*}$, \dots) and user-defined function symbols build terms, and comparison symbols ($\texttt{=}$, $\texttt{>=}$, \dots) relate two terms to form atomic formulae. The first definition in Figure~\ref{fig:query} uses the connectives and a quantifier: $\mathtt{Snip}(g,s)$ holds when \emph{some} member of group $g$ produced snippet $s$ and $s$ matches the topical regexp.

\paragraph{Temporal operators.} Temporal operators relate different
time-points using three operators present in \sysname. $\Prev\,\phi$ holds at $i$ if $i>1$ and $\phi$ held at the previous time-point $i-1$. $\Once\,\phi$ holds at $i$ if $\phi$ held at some time-point $j \leq i$.  $\psi \Since \phi$ holds at $i$ if $\phi$ held at some $j \leq i$ and $\psi$ held at every time-point after $j$, up to and including $i$. An optional
interval $I$ bounds how far in the past $\phi$ may have held:
$\Once[0,180]\,\phi$ reads ``$\phi$ held at some point within the last $180$ days'', i.e.\ $\tau_i - \tau_j \in [0,180]$. We use this operator to define the rolling window in our running example. Notice that two granularities of time are involved: $\Prev$ steps back one position in the trace, whereas intervals such as $[0,180]$ are measured in timestamp units and define intervals. We restrict ourselves to past-only operators because they admit incremental streaming evaluation~\cite{Basin2015}, which our symbolic engine relies on.

\paragraph{Aggregations and $\Let$.} Deciding whether a stance is
\emph{mostly} positive is a quantitative question, answered by aggregations. An aggregation $\agg{y}{\color{black}\omega}{t}{x_1,\dots,x_m}{\phi}$ binds $y$ to the aggregate $\omega \in \{\kw{CNT}, \kw{MIN}, \kw{MAX}, \kw{AVG}, \kw{MED}, \kw{SUM}\}$ of the term $t$ (first position) over all satisfactions of the formula $\phi$ (third position), grouped by the variables $x_i$ (second position), similar to SQL's \texttt{GROUP~BY}~\cite{Basin2015b}. Finally, $\Let$ introduces named intermediate predicates, so that complex queries can be written modularly and as a combination of simple definitions and queries, enabling easier usage and better overview of the written specification. We can see the usage of these operators in the running example with $\mathtt{PosN}(g,x)$: the body uses the built-in atom $\mathtt{tp}(t)$, which binds $t$ to the index of the current time-point. This means that the body holds for the indices of the time-points within the last $180$ days at which group $g$ produced a positive snippet.  $\kw{CNT}$ then binds $x$ to the number of such indices. 

\paragraph{The complete query.} We can look at the example specification shown in Figure~\ref{fig:query} and follow how \sysname assembles these pieces
to be able to analyze a large corpus of political speeches. $\mathtt{Snip}$ is a regexp pre-filter that discards off-topic speeches that are not relevant to our analysis and the next two $\Let$s classify the stance of the \emph{already filtered} snippets with LLM atoms, so the number of model calls is bounded by the filtered set rather than by the corpus size. Two $\Once[0,180]$ aggregations count the recent time-points with positive and with negative snippets per group, and $\mathtt{MostlyPos}$ compares the counts. Because a grouped aggregation yields no value for an empty group, $\mathtt{MostlyPos}(g)$ holds only when the window contains at least one snippet of either positive or negative sentiment. The final line of the query fires when a group produces a negative snippet, was
mostly positive at the previous time-point, and is no longer mostly positive
now. %One consequence is that the query cannot fire at a group's first negative snippet: at that point, $\mathtt{MostlyPos}(g)$ cannot have held at the previous time-point, because the window then contained no negative snippet and hence no $\mathtt{NegN}$ value. 
Appendix~\ref{sec:app:queries} lists the full queries used in our
experiments.

\begin{figure}[t]
\centering
\small
\setlength{\arraycolsep}{1.5pt}
$\begin{array}{r@{\;}c@{\;}l@{}}
t      &::=& c \mid x \mid f(t,\dots,t)\\
\omega &::=& \kw{CNT} \mid \kw{MIN} \mid \kw{MAX} \mid \kw{AVG} \mid \kw{MED} \mid \kw{SUM}\\
I      &::=& [i,i] \mid [i,*]\\
\phi   &::=& e(t,\dots,t) \mid \Not\phi \mid \phi \And \psi \mid \phi \Or \psi \mid \Exists x. \phi \\
       &\mid&
       \, \Prev I?\,\phi \mid \Once I?\,\phi \mid \phi \Since I? \psi\ \\
       &\mid& \Let e(x,\dots,x) = \phi \In \psi \mid \agg{x}{\omega}{t}{x,\dots,x}{\phi} \\
       &\mid& t \bowtie t \mid \mathtt{tp}(t) \mid \prompt{prompt(x,\dots,x)} \mid \regexp{regexp}{x}
\end{array}$
\caption{Syntax of the \sysname query language. Here $e$ ranges over event
and defined predicate names, $c$ over constants, $x$ over variables, $f$ over
function symbols, $\psi$ over formulae (like $\phi$), $i$ over natural
numbers (in timestamp units, e.g.\ days), and
${\bowtie} \in \{\texttt{=}, \texttt{<}, \texttt{<=}, \texttt{>},
\texttt{>=}\}$; $I?$ denotes an optional interval, and $\mathtt{tp}(t)$ binds
$t$ to the index of the current
time-point.\label{fig:query-language}}
\end{figure}

\subsection{Query Engine}\label{sec:engine}

A na\" ive approach to evaluating Natural-language predicates would traverse the formula top-down, calling the LLM every time the logical evaluation
reaches a natural-language atom. However, this approach is slow because each model call blocks the evaluation, repeats identical calls, and increases the tool's runtime cost. To avoid this issue, the
\sysname engine evaluates a query in four stages:
\textbf{(i)} compilation of the surface query into a pure-MFOTL formula plus
auxiliary Python predicates; \textbf{(ii)} extraction of the non-temporal
$\Let$ prefix of the query; \textbf{(iii)} parallel pre-evaluation of that
prefix over all time-points; and \textbf{(iv)} streaming the resulting events
into a symbolic monitor that evaluates the temporal part of the query.
Figure~\ref{fig:pipeline} shows the resulting logical data flow on the query
of Figure~\ref{fig:query}: the panels follow the query's cascade, with stage
(iii) producing the second and third panels and stage (iv) the last one. Our
symbolic backend is EnfGuard~\cite{Hublet2025}, a state-of-the-art monitoring
and an enforcement engine for MFOTL that takes an MFOTL formula and a Python file containing function definitions as input.

\paragraph{(i) Compilation.} As the symbolic backend of \sysname does not understand either natural language prompts or regexps, these expressions are compiled into ordinary function symbols. Each distinct natural-language atom $\prompt{T}$ is assigned a fresh function symbol $\mathtt{llm\_pred}_k$ and replaced by
$\mathtt{llm\_pred}_k(x_1,\dots,x_m) = 1$, where the $x_i$ are $T$'s
placeholders. The compiler binds a Python function to each natural-language atom that calls the configured LLM backend on the instantiated prompt. In our example, the two stance atoms become two such functions. Likewise, regexp atoms compile into
$\mathtt{re\_pred}_k$, wrapping \texttt{re.search}. Repeated templates are
de-duplicated into one function, which enables memoization.

\paragraph{(ii) Non-temporal $\Let$ prefix extraction.} Invoking the LLM from
\emph{inside} the monitor would block it on every model call and force serial
evaluation. However, the leading $\Let$s of a query are typically
non-temporal, per-document filters and do not strictly require sequential evaluation. For example, in Figure~\ref{fig:query},
$\mathtt{Snip}$, $\mathtt{Pos}$, and $\mathtt{Neg}$ only look at the current
time-point. The engine therefore strips the maximal initial run of $\Let$s
whose bodies contain no temporal operator, leaving a reduced formula in which
they appear as base predicates.

\paragraph{(iii) Parallel pre-evaluation.} Because the stripped $\Let$s are
non-temporal, they can be evaluated on each time-point independently, and
hence in parallel. The engine analyses them for mutual dependencies (in our
example, $\mathtt{Pos}$ and $\mathtt{Neg}$ depend on $\mathtt{Snip}$) and
evaluates them level by level against each local database $D_i$, dispatching
independent $\Let$s concurrently. For the same reason, this stage can also
run incrementally as new documents arrive. The backend exposes a thread-safe memo cache, so each prompt incurs only a single model call instead of multiple.

\paragraph{(iv) Streaming symbolic evaluation.} Finally, EnfGuard runs the
reduced formula (in our example, the rolling counts, their comparison, and
the shift condition) over the trace extended with the precomputed $\Let$
events. EnfGuard returns satisfying valuations per time-point. The
engine joins them back to the predicate-level decisions to report the evidence behind each match: for $(\ast)$, the shifting
group, the time-point of the shift, and the classified snippets behind the counts. EnfGuard evaluates a well-defined, expressive syntactic fragment of MFOTL and rejects formulae outside it. All queries in this paper lie within this fragment.
\section{Evaluation}
\label{sec:eval}

Our experiments test whether using TL;QR improves accuracy and cost over prompting an LLM directly. Concretely, we ask three questions. \textbf{RQ1:} Is \sysname more accurate than prompt-based baselines on temporally complex queries? \textbf{RQ2:} How
much of the baseline error comes from temporal reasoning rather than annotation? \textbf{RQ3:} How do accuracy and cost depend on the underlying LLM?

\subsection{Experimental Setup}
\label{sec:setup}

\paragraph{Methods.} We compare \sysname to four baselines: \textsc{Key-words}, \textsc{Zero-shot}, \textsc{few-shot} and $\text{LLM}^{\text{LLM}}$. \textsc{Key-words} is a purely symbolic baseline that
pairs a hand-written lexicon with a hand-written temporal heuristic and
makes no LLM calls.
\textsc{Zero-shot} and \textsc{few-shot} LLMs read the raw corpus in
chronological batches and output matches directly.
$\text{LLM}^{\text{LLM}}$ is a stronger, diagnostic baseline: a first LLM
extracts the same atoms that \sysname uses, and a second LLM then reasons
temporally over the extracted atom stream. Since extraction is shared, any
gap between $\text{LLM}^{\text{LLM}}$ and \sysname is attributable to
the temporal reasoning provided by \sysname's symbolic component. All LLM-based methods run with all three \texttt{gpt-5.6} variants: \texttt{luna}, \texttt{terra}, and \texttt{sol}.

\paragraph{Metrics.} A predicted match is a (group, timestamp) pair. We
report F1 over matches that fall within a fixed tolerance of the true
timestamp; for \textsc{stance shift}, the reference point is the time at
which the shift becomes detectable, half a window after it begins. For
localisation quality, we report MAE, the Mean Absolute Error (offset in 
timestamps) between a reported match and the planted event. For cost we
count corpus passes weighted by the relative input-token price of the
model, in units of one zero-shot pass with the cheapest model (\texttt{luna}); a query with
$k$ LLM atoms costs $k$ passes. All numbers are means over three random
seeds.

\paragraph{Benchmarks.} We construct three synthetic parliamentary
benchmarks, each with its own data-generating process, event signature,
query, and planted ground truth; the full queries are given in
Appendix~\ref{sec:app:queries}. \textsc{Stance shift} follows groups debating immigration (similar to Figure~\ref{fig:query}), and a group is reported when it turns from mostly positive to mostly negative, where ``mostly'' compares rolling counts of positive and negative speeches over a window of
$20$ time-points. \textsc{Unresolved hedge} targets broken commitments: a
speaker is reported when they postponed their position on a proposal between
$5$ and $15$ time-points ago and have never committed to that proposal since, so
the query combines a metric deadline with the absence of a resolving event.
\textsc{Coordinated pile-on} is concerned with concentrated hostility: a target is
reported when it accumulates at least $15$ hostile claims within a
$20$-time-point window and exceeds every other target's count by a margin of
$4$, with the query comparing aggregates across all actors. Together, the three
benchmarks exercise rolling counts, metric deadlines, and cross-group
comparison. The generator plants both positive and negative instances, and
the corpus text is LLM-written from per-class specifications and cached, so
all methods and models see identical corpora. Each benchmark comes in three
sizes ($1.5$k, $3$k, and $4.5$k events), obtained by scaling the number of
actors with fixed time spans and detection windows.

\subsection{Results}

% ---- preamble (or just above the table) ----
% F1 / MAE in one cell, both at table size.
\newcommand{\fm}[2]{#1\,\slash\,#2}
% Recall = 0: MAE undefined. Phantom keeps column widths even.
\newcommand{\fmn}[1]{#1\,\phantom{\slash\,0.0}}

\begin{table}[th!]
  \centering
  \small
  \setlength{\tabcolsep}{3pt}
  \begin{tabular}{@{}ll ccc c@{}}
  \toprule
  & & \multicolumn{3}{c@{}}{\textbf{F1}$\uparrow$ \slash\ \textbf{MAE}$\downarrow$} \\
  \cmidrule(l){3-5}
  \textbf{Method} & \textbf{LLM} & 1.5k & 3k & 4.5k & \textbf{Cost} \\
  \midrule
  \multicolumn{5}{@{}l@{}}{\textsc{Stance shift}} \\
  \midrule
  0-shot             & \texttt{luna}  & \fm{\textbf{100.0}}{0.0}          & \fm{93.3}{2.2}           & \fm{60.0}{0.0} & 1$\times$ \\
                     & \texttt{sol}   & \fm{\textbf{100.0}}{0.0} & \fm{93.3}{0.0}           & \fm{\textbf{100.0}}{0.0} & 25$\times$ \\
  \addlinespace[1.5pt]
  5-shot             & \texttt{luna}   & \fm{\textbf{100.0}}{0.0}          & \fm{100.0}{2.2}          & \fm{93.3}{0.0} & 1$\times$ \\
                     & \texttt{sol}   & \fm{\textbf{100.0}}{0.0}          & \fm{\textbf{100.0}}{0.0}          & \fm{\textbf{100.0}}{0.0} & 25$\times$ \\
  \addlinespace[1.5pt]
  LLM$^{\text{LLM}}$ & \texttt{luna}   & \fm{26.7}{0.0}           & \fm{43.3}{0.0}           & \fm{86.7}{0.0} & 1$\times$ \\
                     & \texttt{terra} & \fm{16.7}{19.0}          & \fm{53.3}{1.5}           & \fm{53.3}{1.5} & 10$\times$ \\
                     & \texttt{sol}   & \fmn{0.0$^{\dagger}$}    & \fm{66.7}{0.0}           & \fm{33.3}{0.0} & 25$\times$\\
  \addlinespace[1.5pt]
  Key-words          & ---   & \fmn{0.0}                & \fmn{0.0}                & \fmn{0.0} \\
  \addlinespace[1.5pt]
  \sysname           & \texttt{luna}   & \fm{\textbf{100.0}}{0.0}          & \fm{\textbf{100.0}}{0.0}          & \fm{\textbf{100.0}}{0.0} & 2$\times$ \\
                     & \texttt{terra} & \fm{\textbf{100.0}}{0.0}          & \fm{\textbf{100.0}}{0.0} & \fm{\textbf{100.0}}{0.0} & 20$\times$ \\
                     & \texttt{sol}   & \fm{\textbf{100.0}}{0.0}          & \fm{\textbf{100.0}}{0.0}          & \fm{\textbf{100.0}}{0.0} & 50$\times$ \\
  \midrule
  \multicolumn{5}{@{}l@{}}{\textsc{Unresolved hedge}} \\
  \midrule
  0-shot             & \texttt{luna}   & \fm{35.6}{2.2}           & \fm{29.6}{4.0}           & \fm{27.0}{3.0} & 1$\times$  \\
                     & \texttt{sol}   & \fm{35.6}{0.0}           & \fm{29.6}{0.0}           & \fm{45.1}{0.0} & 25$\times$  \\
  \addlinespace[1.5pt]
  5-shot             & \texttt{luna}   & \fm{31.7}{0.0}           & \fm{14.8}{0.0}           & \fm{34.8}{0.0}   & 1$\times$ \\
                     & \texttt{sol}   & \fm{35.6}{0.0}           & \fm{29.6}{0.0}           & \fm{45.1}{0.0}  & 25$\times$  \\
  \addlinespace[1.5pt]
  LLM$^{\text{LLM}}$ & \texttt{luna}   & \fm{11.1}{0.0}           & \fm{12.0}{2.2}           & \fm{15.6}{2.0}  & 1$\times$  \\
                     & \texttt{terra} & \fm{9.5}{9.0}            & \fm{4.2}{0.0}            & \fm{14.4}{1.1}  & 10$\times$  \\
                     & \texttt{sol}   & \fmn{0.0}                & \fm{4.2}{1.0}            & \fm{5.8}{1.0} & 25$\times$  \\
  \addlinespace[1.5pt]
  Key-words          & ---   & \fm{22.2}{0.0}           & \fm{13.3}{0.0}           & \fm{14.3}{0.0} \\
  \addlinespace[1.5pt]
  \sysname           & \texttt{luna}   & \fm{64.4}{2.6}           & \fm{\textbf{75.4}}{2.0}  & \fm{\textbf{73.7}}{1.2}  & 2$\times$  \\
                     & \texttt{terra} & \fm{62.6}{2.1}           & \fm{67.6}{1.6}           & \fm{68.6}{0.7}  & 20$\times$  \\
                     & \texttt{sol}   & \fm{\textbf{66.7}}{2.1}  & \fm{69.6}{0.8}           & \fm{68.6}{0.1}  & 50$\times$  \\
  \midrule
  \multicolumn{5}{@{}l@{}}{\textsc{Coordinated pile-on}} \\
  \midrule
  0-shot             & \texttt{luna}   & \fmn{0.0$^{\dagger}$}    & \fmn{0.0}                & \fmn{0.0}  & 1$\times$  \\
                     & \texttt{sol}   & \fm{16.7}{13.0}          & \fm{33.3}{10.5}          & \fmn{0.0$^{\dagger}$}  & 25$\times$  \\
  \addlinespace[1.5pt]
  5-shot             & \texttt{luna}   & \fmn{0.0$^{\dagger}$}    & \fmn{0.0$^{\dagger}$}    & \fmn{0.0$^{\dagger}$}  & 1$\times$  \\
                     & \texttt{sol}   & \fm{33.3}{12.0}          & \fm{33.3}{11.5}          & \fm{16.7}{11.0}  & 25$\times$\\
  \addlinespace[1.5pt]
  LLM$^{\text{LLM}}$ & \texttt{luna}   & \fm{16.7}{10.0}          & \fmn{0.0$^{\dagger}$}    & \fmn{0.0}  & 1$\times$  \\
                     & \texttt{terra} & \fm{16.7}{10.0}          & \fmn{0.0$^{\dagger}$}    & \fmn{0.0$^{\dagger}$}  & 10$\times$  \\
                     & \texttt{sol}   & \fm{33.3}{14.3}          & \fmn{0.0$^{\dagger}$}    & \fmn{0.0$^{\dagger}$}  & 25$\times$  \\
  \addlinespace[1.5pt]
  Key-words          & ---   & \fmn{0.0$^{\dagger}$}    & \fmn{0.0}                & \fm{16.7}{4.0} \\
  \addlinespace[1.5pt]
  \sysname           & \texttt{luna}   & \fm{\textbf{100.0}}{0.0}          & \fm{\textbf{100.0}}{0.0}           & \fm{100.0}{0.1} & 1$\times$  \\
                     & \texttt{terra} & \fm{\textbf{100.0}}{0.0} & \fm{\textbf{95.2}}{0.0}  & \fm{\textbf{100.0}}{0.0}  & 10$\times$  \\
                     & \texttt{sol}   & \fm{\textbf{100.0}}{0.0}          & \fm{\textbf{100.0}}{0.0}           & \fm{\textbf{100.0}}{0.0}  & 25$\times$  \\
  \bottomrule
  \end{tabular}
  \caption{Full results on the three synthetic benchmarks: F1 (\%) and MAE
  (in timestamps), separated by a slash, mean over 3 seeds, for corpus
  sizes $n \in \{1.5\mathrm{k}, 3\mathrm{k}, 4.5\mathrm{k}\}$. %Models are abbreviated:
  %\texttt{luna}, \texttt{terra}, and \texttt{sol} stand for
  %\texttt{gpt-5.6-luna}, \texttt{gpt-5.6-terra}, and \texttt{gpt-5.6-sol}.
  Bold: best F1 per column/benchmark. %\textsc{Key-words} uses no
  $^{\dagger}$Recall $=0$: the
  method localised no true instance; precision is undefined, F1
  collapses and MAE is omitted. %The cost multiplier is based on input token cost at list prices, neglecting prompt length, with the cost of a full LLM pass on the dataset with the cheapest model (\texttt{luna}) set to 1.
  }
  \label{tab:results-full}
\end{table}

\paragraph{Accuracy (RQ1).} Table~\ref{tab:results-full} shows the full
results. \sysname obtains the best F1 on all three benchmarks, with the
largest margins where temporal structure is non-trivial: $62$--$75$\% vs.
$30$--$45$\% on \textsc{unresolved hedge} and $95$--$100$\% vs. $17$--$33$\% on
\textsc{coordinated pile-on}. On \textsc{Stance shift},~few-shot prompting also reaches $100$\%, so the
benchmark is saturated at this scale. It discriminates between methods only
under weaker models (Table~\ref{tab:results-full}).
\
\paragraph{Temporal reasoning versus extraction (RQ2).}
$\text{LLM}^{\text{LLM}}$ consumes the same atoms as \sysname, yet it reaches
only $33.3$, $3.3$, and $11.1$ F1 on the three benchmarks. This gap, therefore, isolates temporal reasoning from extraction, showing the expected LLM weaknesses at counting and ordering over longer time horizons. 
The results of the symbolic \textsc{key-words} baseline are generally poor, with some findings on \textsc{hedge}, whose phrasings are lexically
distinctive, but none on \textsc{stance}, where polarity must be read
through negation and attribution that a marker lexicon cannot resolve. MAE
further illustrates this: \sysname localizes stance shifts and pile-ons exactly ($0.0$) and hedges to within $0.1$--$2.6$ timestamps, whereas every LLM-only baseline that finds a pile-on is off by $10$--$14$ timestamps.

\paragraph{Model sensitivity and cost (RQ3).} Across the three backends,
\sysname varies by at most $5$ F1 and is identical on two benchmarks, while
the baselines swing widely (Table~\ref{tab:results-full}). \sysname's performance is thus generally robust to the scale of models used for the natural language queries. This robustness to model size further supports the use of cheaper models, further reducing the costs of corpus analysis with \sysname. The last column of Table~\ref{tab:results-full} prices each configuration in units of a single zero-shot pass over the corpus using the cheapest model. Since \sysname keeps its accuracy with the cheapest backend,
\textsc{stance shift} costs $2$ units for a perfect score, against $25$
units for zero-shot prompting on \texttt{sol}.
Our approach provides better or equivalent accuracy at a much lower cost.

\section{Case Study: GDPR stances in the European Parliament}
\label{sec:case}

To test \sysname{} on real-world data, we answer an open question relevant to both political science and privacy studies: \emph{did any parliamentary group's stance on the General Data
Protection Regulation (GDPR) in the European Parliament shift from supportive or neutral to critical over time, and when?}

\paragraph{Setup and query.} We use the European Parliament section of
ParlLawSpeech~\cite{ParlLawSpeech} from 2010 onwards: $442{,}593$ multilingual plenary speeches by twelve political groups, covering most of the GDPR's parliamentary history from the 2012 proposal through
2016 adoption to post-2018 application. One formula describes the pattern in four stages. Firstly, a regexp atom matches only the speeches that name the regulation in any official language (GDPR, DSGVO, RGPD, \dots). Afterward, a natural-language atom retains only those genuinely discussing data protection, filtering out those that only use the acronym in passing or in unrelated contexts. The remaining two steps classify supportive versus critical stances, and temporal operators detect the first point at which a group's negative count overtakes its positive count within a rolling $180$-day window. \texttt{gpt-5.6-luna} was used as the LLM backend.
The query's structure is thus similar to Figure~\ref{fig:query}.
The full query and prompts are shown in Appendix~\ref{sec:app:queries}.

\begin{table}[t]
  \centering
  \small
  \setlength{\tabcolsep}{2pt}
  \begin{tabular}{@{}lr rr rrrrr@{}}
  \toprule
  & & \multicolumn{2}{c}{\textbf{IAA}} & \multicolumn{5}{c}{\texttt{gpt-5.6-luna}} \textbf{vs. human} \\
  \cmidrule(lr){3-4} \cmidrule(lr){5-9} 
  \textbf{Predicate} & $n$ & Agr.\ & $\kappa$ & Acc.\ & P & R & F1 & $\kappa$ \\
  \midrule
  Relevance & 50 & 100 & 1.0 & 96 & 1.0 & .937 & .967 & .920 \\
  Positive stance & 50 & 80 & .598 & 75 & .720 & .958 & .822 &.600 \\
  Negative stance & 50 & 96 & .913 & 96 & .680 & 1.0 & .810 & .680  \\
  \bottomrule
  \end{tabular}
  \caption{Human validation of the LLM predicates. Inter-annotator agreement (IAA) columns report raw agreement and Cohen's $\kappa$ between the two primary annotators. Accuracy, precision, recall, F1, and Cohen's $\kappa$ are computed with respect to the adjudicated gold labels on the stratified evaluation and reweighted based on class sizes. % Agreement and accuracy are in \%.
  }
  \label{tab:validation}
\end{table}
\paragraph{Pre-filtering.} The filtering makes the query affordable: the
regexp reduces $442{,}593$ speeches to $343$ potential GDPR mentions.
Next, the relevance atom identifies $186$ speeches that genuinely discuss
the regulation; only those reach the
stance classifier. An LLM-only pipeline would read all $442{,}593$ speeches, whereas our filtering reduces this by three orders of magnitude and can be expressed as part of the query without the need for an external filtering mechanism.

\paragraph{Cost and runtime.} The analysis issues $715$ LLM calls  ($343$ relevance judgments plus two stance atoms for each of the $186$ survivors) in approximately 15 minutes and at a cost of $\sim$\$1. Most computation time is  spent on regexp evaluation and
model calls at $16$-way concurrency.

\paragraph{Findings.} Of the $186$ relevant speeches, $147$ are classified as supportive and
$26$ as critical. The query reveals five shifts by the center-left Greens/EFA in December~2019, the center-right EPP in April~2016 and October~2020,
and the national-conservative ECR in February~2020 and May 2022 (see Figure~\ref{fig:gdpr}). 
We can correlate these shifts with real-life events, with the first EPP shift following the Council's adoption of the text and subsequent shifts occurring during the regulation's implementation phase. Three groups with a supportive stance, the GUE/NGL, S\&D, and Renew groups, and a far-right group with a mixed attitude, the ID group, register no shift.
Each verdict produced by \sysname carries classified speeches as witnesses, so the explanation of how each verdict was made can be inspected.
After inspecting the evidence, the potential Greens/EFA 
shift---a single speaker voicing doubts about 
implementation effectiveness---does not appear to mark a qualitative
departure from the group's overall supportive stance.
In contrast, all other potential shifts feature sustained criticism 
of the law. One also notes shares of critical statements approaching
or touching 50\% by the Renew and EPP groups on several occasions, 
revealing a mixed attitude of center-right members especially in
the immediate aftermath of GDPR's entry into force. 

\paragraph{Human Validation.} In Table~\ref{tab:validation}, we validate the accuracy of our natural-language atoms against human annotations. For each predicate, 50 speeches were selected by stratified random sampling %from the logged predictions 
with balanced numbers of positive and negative predictions. Two authors independently labeled all sampled speeches, with disagreements resolved by a third. The adjudicated labels constitute the gold standard. We find weighted F1 score and recall above $0.8$ and $0.9$, respectively, for \texttt{gpt-5.6-luna}. Since recall is high, expert review of positive annotations can be used at a relatively low cost to further reduce potential inaccuracies.

\section{Conclusion}
We have presented \sysname, a declarative query language and open-source engine for querying temporal discursive patterns in large timestamped corpora. The core novelty of our system is its complementary use of neural and symbolic tools. An LLM is used as a judge that only processes local predicates about individual pieces of text, while a symbolic engine with a formal semantics performs all counting, ordering, and windowing. Our experiments show that \sysname is more accurate than LLM-only baselines and that its advantage is due to exact temporal reasoning. Additionally, \sysname provides good accuracy even with an unexpensive LLM backend, whereas the baselines may degrade sharply.  

Future work will extend support to future operators and richer atoms, such as retrieval-based or cross-document predicates. Currently, \sysname treats every LLM judgment as a Boolean. Since the accuracy of each atom can be measured on a labeled sample, a natural extension is to propagate these measurements through the formula, so that every verdict carries a calibrated confidence instead of a binary truth value.

\section{Limitations}

\paragraph{Atom quality bounds performance.} \sysname's temporal layer is
exact, but its verdicts are only as reliable as the LLM judgments they rest
on. We can observe this on our benchmarks, where, when the monitor is given the ground-truth atoms, it recovers the planted pattern perfectly, so the loss
that \sysname does incur in Table~\ref{tab:results-full} comes entirely from
atom errors. Moreover, because queries are temporal, a single atom error can
affect verdicts at later time-points, for instance, by starting or ending a
rolling majority too early.

\paragraph{Query and prompt authoring.} Writing a \sysname query requires
familiarity with structured query languages, roughly at the level of a
proficient SQL user, together with some comfort with the temporal operators.
We provide a reusable four-step recipe, but we have not run a user study, and
\sysname offers no automated translation from a natural-language description
into a query. The natural-language atoms themselves must also be phrased with
care: the case-study prompts in Appendix~\ref{sec:app:queries} each run to a
paragraph.

\paragraph{Synthetic benchmarks.} Our corpora are generated from
per-class specifications, so their text is less varied than real parliamentary speech and their ground truth is correct by construction rather than by annotation. Absolute scores should therefore be read as relative comparisons between methods under a controlled generator, not as estimates of performance on real-world tasks.

\paragraph{The case study is exploratory.} The GDPR analysis reports
\emph{candidate} shifts from a single corpus and a single query. The
validation in Table~\ref{tab:validation} shows high recall but only moderate
precision for the two stance atoms, so individual classifications require
expert review before any conclusion about European politics is drawn; the
Greens/EFA verdict, which rests on a single speaker and does not survive
inspection, illustrates why. In addition, the regexp pre-filter keeps only
speeches that name the regulation explicitly, so speeches that discuss it in
other words are missed silently, and the cost advantage over a prompt-only
pipeline is an estimate based on token prices.

\paragraph{Engine constraints.} The symbolic backend~\cite{Hublet2025}
evaluates a syntactic fragment of MFOTL: only past-time temporal operators are supported, and some formulae, such as negated operands under temporal
operators, are approximated rather than decided exactly and must therefore be avoided.
Queries outside this fragment have to be rewritten, as the negation-free form of the unresolved-hedge query in Appendix~\ref{sec:app:queries} illustrates.

\section{Ethical considerations}

Incorrect classification of public speeches can misrepresent the
political stance of individuals. 
Since the findings of our case study are only
preliminary, we do not discuss individual speakers' detected stances
and report only group-level outcomes, underlining the need for
further qualitative analysis to reach substantive conclusions.

\section{Licence for artifacts}

EnfGuard is released under the GNU GPLv3 licence. The ParlLawSpeech dataset 
is released under the CC BY 4.0 licence. We release our own software 
under GNU GPLv3.

\section{Acknowledgments} This work was partially supported by the European Research Council (ERC) Grant HYPER (No.~101055412). Views and opinions expressed are, however, those of the authors only and do not necessarily reflect those of the European Union or the European Research Council Executive Agency. Neither the European Union nor the granting authority can be held responsible for them.

\bibliography{custom}

\appendix

\section{MFOTL semantics}
\label{sec:semantics}

Let us fix a trace $\sigma = (\tau_1,D_1),\dots,(\tau_n,D_n)$ and  interpretations
$c^{\mathcal I}$ and $f^{\mathcal I}$ of all constant and function symbols, 
as well as a Boolean prompt evaluator $\mathrm{LLM}$.
Temporal operators without an interval attached are treated as carrying the interval $[0,\infty)$.
The semantics of \sysname{} are given in Figure~\ref{fig:semantics}.

\begin{figure}[t]
  \centering
  \noindent\resizebox{\linewidth}{!}{$\displaystyle
\begin{aligned}
v,i \models e(t_1,\dots,t_k)
 &\iff e(\llbracket t_1\rrbracket_v,\dots,\llbracket t_k\rrbracket_v)\in D_i \\[3pt]
v,i \models t_1 \bowtie t_2
 &\iff \llbracket t_1 \rrbracket_v \bowtie \llbracket t_2 \rrbracket_v \\[3pt]
v,i \models \mathtt{tp}(t)
 &\iff \llbracket t \rrbracket_v = i \\[3pt]
v,i \models \Not \phi
 &\iff v,i \not\models \phi \\[3pt]
v,i \models \phi \And \psi
 &\iff v,i \models \phi \ \land\ v,i \models \psi \\[3pt]
v,i \models \phi \Or \psi
 &\iff v,i \models \phi \ \lor\ v,i \models \psi \\[3pt]
v,i \models \Exists\,x.\,\phi
 &\iff \exists d.\ v[x\mapsto d],i \models \phi \\[4pt]
v,i \models \Prev{I}\,\phi
 &\iff i>1 \land \tau_i-\tau_{i-1}\in I \land v,i-1 \models \phi \\[3pt]
v,i \models \Once{\,I}\,\phi
 &\iff \exists j\le i.\ \tau_i-\tau_j\in I \land v,j \models \phi \\[3pt]
v,i \models \phi \Since{I}\,\psi
 &\iff \exists j\le i.\ \tau_i-\tau_j\in I \land v,j\models\psi \\
 &\qquad\land\ \forall k\,(j<k\le i \Rightarrow v,k\models\phi) \\[4pt]
v,i \models \Let e(\bar x)=\phi \In \psi
 &\iff v,i \models_{\sigma'} \psi \ \text{ where} \\[-1pt]
 &\hspace{-5em} \sigma' = \sigma[e(\bar d) \in D_j \iff
          v[\bar x\mapsto\bar d],j \models \phi] \\[4pt]
v,i \models \agg{x}{\omega}{t}{\bar y}{\phi}
 &\iff v(x)\in\omega(M)\ \text{ where } \bar z = \fv(\phi)\setminus\bar y\  \\[-1pt]
 &\hspace{-5em}\text{and } M =
   \bigl[\ \evaluate{t}{v[\bar z\mapsto\bar d]} \mid \\[-1pt]
 &\hspace{-3em}
        v[\bar z\mapsto\bar d],i \models \phi,\
        \bar d \in \Domain^{|\bar z|} \ \bigr] \\[-1pt]
 &\hspace{-5em}\text{and } |\bar y|>0 \Rightarrow M \neq [\,] \\[4pt]
v,i \models \prompt{prompt(\bar x)}
 &\iff \mathrm{LLM}(\prompt{prompt(\bar x)}[\overline{x} \mapsto v(\overline{x})]) \\[3pt]
v,i \models \regexp{r}{x}
 &\iff v(x)\ \text{matches the regexp } \mathtt{\color{green}r} \\[4pt]
\llbracket c \rrbracket_v
 &= c^{\mathcal I},\qquad
\llbracket x \rrbracket_v = v(x), \\[2pt]
\llbracket f(t_1,\dots,t_k) \rrbracket_v
 &= f^{\mathcal I}(\llbracket t_1 \rrbracket_v,\dots,\llbracket t_k \rrbracket_v)
\end{aligned}$}
  \caption{Semantics of \sysname{}. $\sigma$ is the signature interpretation, $\Domain$ the domain, and $\evaluate{t}{v}$ the value of term $t$ under valuation $v$.}
  \label{fig:semantics}
\end{figure}

\section{Queries Used in the Experiments}
\label{sec:app:queries}

We give the four queries evaluated in Sections~\ref{sec:eval}
and~\ref{sec:case}. Natural-language atoms are shown in full for the synthetic
benchmarks; the case-study prompts are abbreviated, as each runs to a
paragraph.

\subsection{Synthetic Benchmarks}

\paragraph{Stance shift.} A group is reported when it speaks negatively, was
mostly positive one time-point earlier, and is mostly negative now, where
``mostly'' compares rolling counts over a window of $20$ time-points.
\begin{lstlisting}[language=MFOTL-NL]
LET Pos(g) = EXISTS mep, text. Speech(mep,g,text)
  AND "{text} expresses a positive, welcoming, or
       humanitarian stance towards immigration" IN
LET Neg(g) = EXISTS mep, text. Speech(mep,g,text)
  AND "{text} expresses a negative, hostile, or
       restrictive stance towards immigration" IN
LET MostlyPos(g) = EXISTS x, y.
  x <- CNT(t; g; ONCE[0,20] (Pos(g) AND tp(t))) AND
  y <- CNT(t; g; ONCE[0,20] (Neg(g) AND tp(t))) AND
  x > y IN
LET MostlyNeg(g) = EXISTS x, y.
  x <- CNT(t; g; ONCE[0,20] (Pos(g) AND tp(t))) AND
  y <- CNT(t; g; ONCE[0,20] (Neg(g) AND tp(t))) AND
  x < y IN
Neg(g) AND (ONCE[1,1] MostlyPos(g))
  AND MostlyNeg(g)
\end{lstlisting}

\paragraph{Unresolved hedge.} A speaker is reported when they hedged on some
proposal between $d{=}5$ and $d{+}k{=}15$ time-points ago and have not
committed to that same proposal anywhere in the lookback.
\begin{lstlisting}[language=MFOTL-NL]
LET Hedge(sp, prop) = EXISTS text.
  Utterance(sp, prop, text) AND
  "{text} contains an explicit statement postponing
   or withholding a position on the proposal" IN
LET Commit(sp, prop) = EXISTS text.
  Utterance(sp, prop, text) AND
  "{text} contains an explicit pledge to support or
   vote for the proposal, in the speaker's own
   voice" IN
EXISTS prop. (ONCE[5,15] Hedge(sp, prop))
  AND (NOT (ONCE[0,15] Commit(sp, prop)))
\end{lstlisting}
This is a negation-free rendering of the pattern, which is naturally stated
with $\Since$. The two coincide on this generator, because a
resolving commitment is always planted within $d$ of its hedge and therefore
falls inside the $[0,d{+}k]$ lookback.

\paragraph{Coordinated pile-on.} A target is reported when its count of hostile
claims over a $20$-time-point window reaches $\theta{=}15$ and exceeds every
other target's count by a margin of $4$.
\begin{lstlisting}[language=MFOTL-NL]
LET Att(tg) = EXISTS sp, text. Claim(sp,tg,text)
  AND "{text} makes a hostile or accusatory claim
       against the person it refers to" IN
LET Cnt(tg, n) =
  n <- CNT(t; tg;
       ONCE[0,20] (Att(tg) AND tp(t))) IN
LET Cnt2(tg2, m) =
  m <- CNT(t; tg2;
       ONCE[0,20] (Att(tg2) AND tp(t))) IN
EXISTS n. Cnt(tg, n) AND n >= 15 AND
  (NOT (EXISTS tg2, m. Cnt2(tg2, m) AND
        (NOT (tg2 = tg)) AND m + 4 > n))
\end{lstlisting}
The margin makes the comparison robust to atom noise: a bare ``no other target
reaches this count'' is decided by a single count, so a few false-positive
atoms on one bystander flip the verdict.

\begin{figure*}[t]
  \centering
  \includegraphics[width=\textwidth]{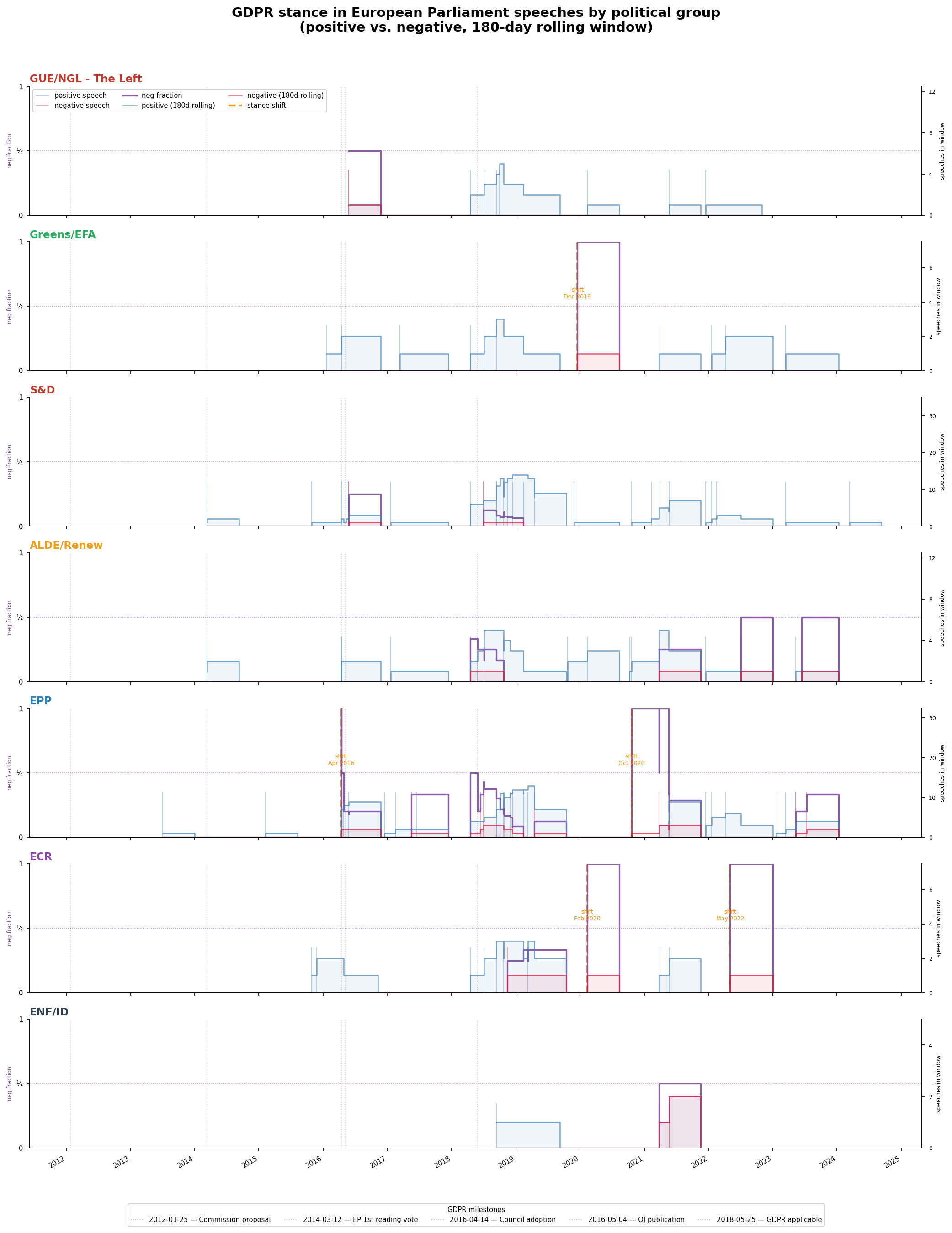}
  \caption{GDPR stance over time by political group, as returned by the
  case-study query. Rug marks are individual supportive (blue) and critical
  (red) speeches; the shaded lines are the $180$-day rolling counts computed by
  the monitor; the purple line is the resulting negative share with its
  one-half threshold; dashed orange lines mark detected stance shifts.
  Dotted grey lines are GDPR milestones. All series are read directly off the
  monitor's output, so every point is traceable to the speeches that produced
  it.}
  \label{fig:gdpr}
\end{figure*}

\subsection{Case Study}

The GDPR query stages a regexp filter, a relevance atom, two stance atoms and a
rolling-window comparison. For readability, the four atom bodies are written
here as the placeholders $\mathtt{GdprRe}$, $\mathtt{P_{rel}}$,
$\mathtt{P_{pos}}$ and $\mathtt{P_{neg}}$; their full texts follow the query.
\begin{lstlisting}[language=MFOTL-NL]
LET Mention(mep,g,text) = Speech(mep,g,text)
  AND /GdprRe/text IN
LET Rel(mep,g,text) = Mention(mep,g,text)
  AND "P_rel" IN
LET Pos(g) = EXISTS mep, text. Rel(mep,g,text)
  AND "P_pos" IN
LET Neg(g) = EXISTS mep, text. Rel(mep,g,text)
  AND "P_neg" IN
LET CntPos(g,x) =
  (x <- CNT(t; g; ONCE[0,180] (Pos(g) AND tp(t))))
  OR (x = 0 AND
      NOT (EXISTS t. ONCE[0,180] Pos(g))) IN
LET CntNeg(g,x) =
  (x <- CNT(t; g; ONCE[0,180] (Neg(g) AND tp(t))))
  OR (x = 0 AND
      NOT (EXISTS t. ONCE[0,180] Neg(g))) IN
LET NegFrac(g,r) = EXISTS p, n.
  (ONCE[0,180] (Pos(g) OR Neg(g)) AND CntPos(g,p)
   AND CntNeg(g,n) AND p + n > 0;
   r <- (1000 * n) @\texttt{/}@ (p + n)) IN
LET MostlyNeg(g) =
  EXISTS r. NegFrac(g,r) AND r > 500 IN
ONCE[0,180] (Pos(g) OR Neg(g)) AND MostlyNeg(g)
  AND PREV (NOT MostlyNeg(g))
\end{lstlisting}
Note the syntax $\phi\kw{;}\,x \gets t$, which is syntactic sugar
for $\agg{x}{\kw{AVG}}{t}{y_1,\dots,y_n}{\phi}$, where $y_1,\dots,y_n$
are the free variables of $\phi$. This behaves exactly like a traditional
assignment.

Timestamps are days, so the $\Once[0,180]$ windows are six calendar months.
$\mathtt{CntPos}$ and $\mathtt{CntNeg}$ carry an explicit zero case so that a
group with no speeches of one polarity still yields a count, and
$\mathtt{NegFrac}$ reports the negative share in per-mille to stay within
integer arithmetic.

\paragraph{$\mathtt{GdprRe}$.} This is a $42$-term alternation over the regulation's
official names and abbreviations in the EU's official languages, e.g.\
\texttt{\textbackslash bGDPR\textbackslash b}, \texttt{\textbackslash bDSGVO\textbackslash b},
\texttt{\textbackslash bDS-GVO\textbackslash b}, \texttt{\textbackslash bRGPD\textbackslash b},
\texttt{\textbackslash bAVG\textbackslash b}, \texttt{\textbackslash bRODO\textbackslash b},
\texttt{Datenschutzgrundverordnung},
\texttt{réglement général de protection des données},
and equivalents in all 24 official EU languages.

\paragraph{$\mathtt{P_{rel}}$ (relevance).}
\begin{quote}\footnotesize
Does the following speech mention the GDPR
(General Data Protection Regulation), identified by one of its official names or abbreviations: GDPR (English), DS-GVO or Datenschutz-Grundverordnung (German), RGPD (French / Spanish / Italian / Portuguese), AVG (Dutch), RODO (Polish), or equivalent official abbreviations in other EU languages? Answer 1 ONLY if the speech explicitly names or abbreviates this specific regulation. Answer 0 if: the speech does not mention the GDPR explicitly.
\texttt{<speech>\{text\}</speech>}
\end{quote}

\paragraph{$\mathtt{P_{pos}}$ (supportive stance).}
\begin{quote}\footnotesize
In the following speech, the speaker's OVERALL or NET attitude towards the GDPR
(General Data Protection Regulation / DS-GVO / RGPD / AVG / RODO) is POSITIVE
or SUPPORTIVE. This includes: praising the regulation as a landmark achievement
or a model for the world; welcoming its adoption, entry into force, or
enforcement; calling for stronger or broader application; defending it against
weakening attempts; framing it as a necessary safeguard for citizens' rights;
expressing satisfaction that it was agreed or achieved; or endorsing its
principles even while acknowledging some complexity or room for improvement.
The speaker need not use the word `GDPR' explicitly --- references to the
underlying regulation by its full name or national abbreviation count.
IMPORTANT: focus only on what the speaker says *about the GDPR itself*, not the
overall topic of the speech. A speech primarily about terrorism, migration, or
AI that merely cites `we must respect the GDPR' expresses no positive stance
about the GDPR. Answer 0 only if the speech is predominantly critical, purely
procedural, or takes no stance. \texttt{<speech>\{text\}</speech>}
\end{quote}

\paragraph{$\mathtt{P_{neg}}$ (critical stance).}
\begin{quote}\footnotesize
In the following speech, the speaker's OVERALL or NET attitude towards the GDPR
(General Data Protection Regulation / DS-GVO / RGPD / AVG / RODO) is NEGATIVE,
CRITICAL, or SCEPTICAL. This includes: criticising the regulation as overly
burdensome, complex, costly, or harmful; calling for its simplification,
derogation, reform, or repeal; framing it as an obstacle to innovation,
research, business, public health, or security; opposing specific enforcement
actions as disproportionate; arguing it hinders cross-border data flows, AI
development, or economic competitiveness; or expressing frustration or
disappointment with how it has been applied or interpreted. The speaker may
acknowledge some merits of the GDPR while still being predominantly critical.
The speaker need not use the word `GDPR' explicitly --- references to the
underlying regulation by its full name or national abbreviation count.
IMPORTANT: focus only on what the speaker says *about the GDPR itself*, not the
overall topic of the speech. A speech primarily about terrorism, migration, or
security that merely says `measures must respect the GDPR' expresses no
negative stance about the GDPR. Answer 0 if: the speech is predominantly
supportive of the GDPR; is purely procedural; takes no stance; or mentions the
GDPR only as a legal safeguard, constraint, or compliance requirement to be
respected --- without actually critiquing it.
\texttt{<speech>\{text\}</speech>}
\end{quote}

\end{document}